\documentclass{article}
\usepackage[utf8]{inputenc}
\pdfoutput=1

    \usepackage[preprint]{neurips_2020}

\usepackage[utf8]{inputenc} % allow utf-8 input
\usepackage[T1]{fontenc}    % use 8-bit T1 fonts
\usepackage{hyperref}       % hyperlinks
\usepackage{url}            % simple URL typesetting
\usepackage{booktabs}       % professional-quality tables
\usepackage{amsfonts}       % blackboard math symbols
\usepackage{nicefrac}       % compact symbols for 1/2, etc.
\usepackage{microtype}      % microtypography

\usepackage{hanging} % hanging indents

\usepackage{graphicx}

\usepackage[ruled,vlined]{algorithm2e}
\usepackage{siunitx}

\usepackage{amsthm}
\usepackage{amsmath}
\usepackage{amssymb}

\usepackage{float}

\newtheorem{definition}{Definition}

\newcommand{\R}{\mathbb{R}}
\newcommand{\M}{\mathcal{M}}

\newcommand{\vol}{\text{vol}}

\newcommand{\der}{\mathrm{d}}

\usepackage{}\newcommand{\Net}{{N}}
\newcommand{\netOneDMAP}{\phi}
\newcommand{\netTwoDMAP}{\psi}

\usepackage[utf8]{inputenc}

\usepackage{xcolor}

\title{Transformations between deep neural networks}

\newcommand{\hopkinsOne}{
Department of Chemical and Biomolecular Engineering
\\
Johns Hopkins University
\\
Baltimore, MD
}

\newcommand{\hopkinsTwo}{
Departments of Chemical and Biomolecular Engineering \& \\
 Applied Mathematics and Statistics
\\
Johns Hopkins University
\\
Baltimore, MD
}

\author{Tom Bertalan \\ \hopkinsOne
\And
Felix Dietrich
\\
Department of Informatics
\\
Technical University of Munich
\\
Munich, Germany % https://foursquare.com/fedicon
\And
Ioannis G. Kevrekidis \\ \hopkinsTwo \\ \texttt{yannisk@jhu.edu}
}

\begin{document}

\maketitle

%\todo{I added back \S\ref{sec:differentInputs} from the SI; please review slightly.}

%\todo{Add acknowledgements section.}

%\todo{Delete some of the commented-out redundant sections.}

\begin{abstract}
We propose to test, and when possible establish, 
an equivalence between two different artificial neural networks 
by attempting to construct a data-driven transformation
between them, using manifold-learning techniques.
In particular, we employ diffusion maps with a Mahalanobis-like metric.
If the construction succeeds, the two networks can be thought of as belonging to the same equivalence class.
We first discuss transformation functions between only the {\em outputs} of the two networks;
we then also consider
transformations that take into account outputs (activations)  of a number of internal neurons from each network.
In general, Whitney's theorem dictates the number of measurements from one of the networks 
required to reconstruct each and every feature of the second network.
The construction of the transformation function relies on a consistent, {\em intrinsic} representation of the 
network input space.
We illustrate our algorithm by matching neural network pairs trained to learn
(a) observations of scalar functions;
(b) observations of two-dimensional vector fields; and 
(c) representations of images of a moving three-dimensional object (a rotating horse).
The construction of such equivalence classes across different network instantiations clearly relates to
transfer learning.
We also expect that it will be valuable in establishing equivalence between
different Machine Learning-based models of the same phenomenon observed through different instruments and by different research groups.
\end{abstract}

% \tableofcontents

\section{Introduction}

In our Big Data era, and as the interactions between physics-based modeling and data-driven modeling intensify, we will be encountering---with increasing frequency---situations where different researchers arrive at different data-driven models of the same phenomenon, even if they use the same training data. 
The same neural network, trained by the same research group, on the same data, with the same choice of inputs and outputs, the same loss function and even the same training algorithm---but just different initial conditions and/or different random seeds, will give different learned networks, even though their outputs may be practically indistinguishable on the training set.
This becomes much more interesting when the same phenomenon is observed through different sensors, and one learns a predictive model of the same phenomenon but based on different input variables---how  will we realize that two very different predictive models of the same phenomenon ``have learned the same thing''? In other words, how can we establish, in a data-driven way that two models are transformations of each other? 
The problem then becomes: if we suspect that two neural networks embody models of the same phenomenon, how do we test this hypothesis? How do we establish the invertible, smooth transformation that embodies this equivalence? And, in a second stage, can we test how far (in input space) this transformation will successfully generalize? 
Our goal is to implement and demonstrate a data-driven approach to establishing ``when two neural networks have learned the same thing'' as our title indicates: the types of observations of the networks that can be used, and an algorithm that allows us to construct and test the transformation between them. Our work is based on nonlinear manifold learning techniques, and, in particular, Diffusion Maps using a Mahalanobis-like metric~\citep{singer-2009,berry-2015,dsilva-2016}. This is an attempt at ``gauge invariant data mining''---processing observations in a way that maps them to an intrinsic description, where the transformation between different models is easy to perform (in our case, it can be a simple, orthogonal transformation between intrinsic descriptions). In effect, we construct data-driven nonlinear observers of each network based on the other network. 
We remind the reader that when we talk about ``observations of the networks'', we do not mean their parameters after training; we mean observations of the ``action'' of the trained networks on input data: (some of) the resulting internal activations.
These observations can be based on either output neurons or internal neurons of the networks, which we assume to be smooth functions over a finite dimensional input manifold. Whitney's theorem then guides us in selecting an upper bound on how many of these functions are needed to preserve the manifold topology, and to embed the input manifold itself~\citep{whitney-1936}.

% Our contribution can be summarized as follows:
%\begin{enumerate}
%    \item We consider both the output and internal neurons of deep neural networks as smooth functions over the input manifold. This allows us to employ Whitney's theorem as an upper bound on how many of these functions are needed to preserve the manifold topology, and to use them to embed the input manifolds.
%    \item We define and implement transformation functions between deep neural networks.
%    \item Using the transformation, we can map from the input of one network, through its internal neurons, to the output of the other network.
%    \item We analyze the generalization properties of networks.
%\end{enumerate}
% use this simple example to demonstrate our idea

\begin{figure}
    \centering
    \includegraphics[width=.5\textwidth]{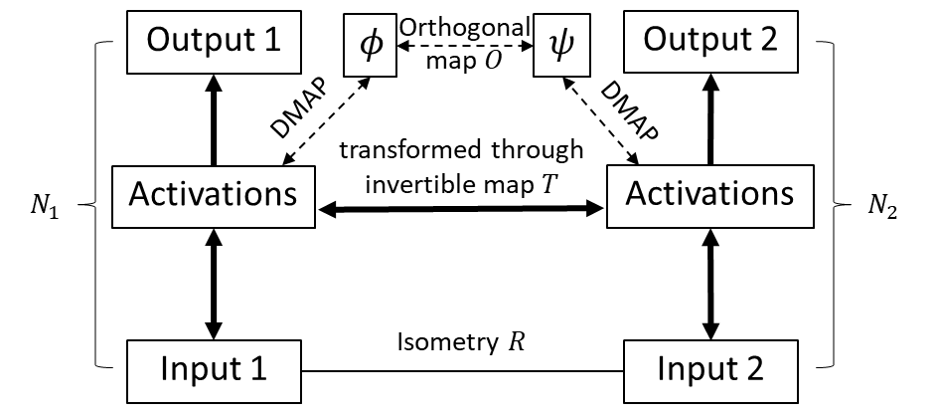}
    \caption{Transformation $T$ between the activation spaces of two neural networks (with, in general, different inputs). The networks $N_1, N_2$ are not necessarily invertible as maps from input to output, but all other maps shown (input isometry $R$, orthogonal map $O$ between intrinsic Mahalanobis Diffusion Map spaces, Mahalanobis Diffusion Maps $\netOneDMAP,\netTwoDMAP$, and the transformation $T$) are invertible.}
    \label{fig:matching_activationspace}
\end{figure}

We now briefly introduce the concept of transformations between neural networks. Details of the approach can be found in section~\ref{sec:transformations} (also see figure~\ref{fig:matching_activationspace}).
In general, we proceed as follows:
{\bf (1)}  Two smooth functions, the {\em tasks}  $f_1$ and $f_2$, are defined on
the same data set (for example, two separate denoising tasks defined on the same set of images). We will also, importantly, discuss cases of tasks defined on different data sets. In that case, when each data set lives on a separate (but assumed diffeomorphic) manifold, we will discuss conditions that enable our construction, turning the diffeomorphism into an isometry.
%%%
%%%IMPORTANT
%%%\todo{The isometry between the two manifolds must not be explicit (e.g. in the ambient space), but in principle it must be possible to obtain it, e.g. through information about the metric distortion. This is discussed in detail in \S\ref{sec:transformations}.};  
{\bf (2)} We separately train two networks to approximate the tasks (for example, two deep, convolutional, denoising auto-encoder networks).
{\bf (3)} Then, we use the output of ``a few'' internal neurons of each of the networks as an embedding space for the input manifold; the minimum number is dictated through Whitney's theorem;
{\bf (4)} Using the Mahalanobis metric (see~\S\ref{sec:constructing transformation} for details) in a Diffusion Map kernel applied to these internal ``intermediate outputs'' yields two {\em intrinsic embeddings} of them that agree up to an orthogonal transformation $O$; 
{\bf (5)} The transformation between activations of the two networks can then be defined through $T=\netTwoDMAP^{-1}\circ O\circ \netOneDMAP$ (see figure~\ref{fig:matching_activationspace}), i.e., we map from the selected activations of network $1$ into (a) the (Mahalanobis) Diffusion Map representation of these activations with $\netOneDMAP$, then (b) map  through an orthogonal transformation $O$ to the Diffusion Map representation of network $2$ activations, and then finally (c) invert the network $2$ Diffusion Map $\netTwoDMAP$ to obtain the activations of network $2$ (see figure~\ref{fig:matching_activationspace}). 
Since the selected neurons are enough to embed the input manifold, all other internal (or final output layer) activations can be then approximated as functions on the embedding. 

\section{Related work}
% This is the standard section of computer science papers where they talk about other work.
% We give a short bibliography leading from vAEs \citep{kingma_auto-encoding_2014,salimans_markov_2015} to GANs \citep{goodfellow_generative_2014} to WGANS \citep{arjovsky_wasserstein_2017}, noting that these have in common a transformation of a "simple" distribution to a target distribution, and then back off to VAEs as a simpler method that is still sufficient for our use.
% 
With the advent of accessible large-scale SIMD processors and GPUs, efficient backpropagation, and the wider adoption of ``twists'' like ReLU activations, dropout, and convolutional architectures, neural networks and AI/ML research in general experienced a ``third spring'' in the early 2000s, which is still going strong today.
While the bulk of the work was in supervised methods,
this boom also led to greater interest in unsupervised learning methods, in particular highly fruitful new work on neural generative modelling.
\cite{kingma_auto-encoding_2014} and later \cite{salimans_markov_2015} began this thread with their work on variational autoencoders (VAEs), which upgrade traditional autoencoders to generative models.
Different autoencoders, trained on observations of the same system, will give very different latent spaces; these can be thought of as different parametrizations of the same manifold.
Variational autoencoders are then a way to ``harmonize'' several different autoencoders, by enforcing the same prior (distribution over the latent variables). 
In \cite{goodfellow_generative_2014}, the ``adversarial learning'' framework was applied to the VAE task, with separate generator and discriminator networks being trained.
This work was significantly extended by \cite{arjovsky_wasserstein_2017}.
%in which \todo{What is the best half-sentence WGAN summary?}
%
% For the purposes of this paper, it is sufficient to consider VAEs, and simpler supervised % learning problems, 
The idea of establishing a correspondence between different learned representations also  motivates our work.
%We see our work as a continuation of the  representation-learning thread. 
Broadly, we construct invertible transformations between different sets of observations arising from activations of different neural networks. When the construction is successful, we can state that ``the networks have learned the same task'' over our data set; we can then explore the generalization of these transformations beyond the training set. 
%comment{This is sort of lame.}
%\comment{
%    We don't actually train any VAEs in this work. The closest we come is perhaps a \textit{denoising} autoencoder for the horse example, in that I augmented with noisy images (whose targets are their coresponding noiseless images.
    %
%    When we talk about how VAEs are "sufficient for our use", I don't want to make it sound like that's what we're taking here as our network of interest.
    %
%    That said, we should try these representation-mapping techniques on these proper generative models, since their theoretical underpinnings, in my qualitative understanding, are all about giving the network the freedom and incentive to learn meaningful internal representations.
%}
%Terms that are related to our work: meta-learning, transfer learning~\citep{weiss-2016}, hyper-parameter optimization, architecture search, ...

The idea of a transformation between systems has been broadly utilized in the manifold learning community following the seminal paper by \cite{singer-2008}.
\cite{berry-2015} discuss the possibility to represent arbitrary diffeomorphisms between manifolds using such a kernel approach.
Both papers employ the diffusion map framework~\citep{coifman-2006}.
Recently, a paper on a neural network approach for such isometries was uploaded to arXiv~\citep{peterfreund-2020}, as an alternative to diffusion maps (see also the works of~\cite{mcqueen-2016b,schwartz-2019} for approximations of isometries).

\section{Transformations between artificial neural networks}\label{sec:transformations}

Here, we introduce mathematical notation, define the problem, and then illustrate the approach through examples of increasing complexity.
We also introduce the concept of using the Mahalanobis metric with Diffusion Maps to construct the transformation, and discuss its use for neural networks.

\subsection{Mathematical notation and problem formulation}

We start by defining two smooth functions, the tasks $f_1:\M\to\R^m$ and $f_2:\M\to\R^m$, with $m\in\mathbb{N}$. The input domain of both tasks is the manifold $\M$. We assume that $(\M,g)$ is a compact, $d$-di\-men\-sio\-nal Riemannian manifold, embedded in Euclidean space $\M\subseteq\mathbb{R}^k$, $k\geq d$, with $g$ the metric induced by the embedding.
In our first example $\M$ is simply $[-1,1]$ and the metric is just the Euclidean metric on $\mathbb{R}$; it is only later, when we want to transform between networks defined on different manifolds, that considerations of the (possibly different) metrics on these manifolds become important, since our Mahalanobis approach is based on neighborhoods of the data points.
%
% \comment{Since this appears to be the definition of $\mu$, I modified the language a little to make that more clear; please review.}
We will then write $\vol_g$ for the volume form on $\M$ induced by $g$, and write $\mu$ for a given, fixed measure on $\M$ (the ``sampling measure''), which we assume to be absolutely continuous with respect to $\vol_g$ and  normalized so that $\mu(\M)=1$. We assume that the measure $\mu$ has a density $\rho:\M\to\R^+$, $\mu=\rho\vol_g$, that is bounded away from zero on our compact manifold $\M$. 
%satisfies $\rho(p)>0 \forall p\in\M$.
That means sampling points through $\mu$ ``covers the manifold well''.
%
% For an invertible, smooth map $R:(\M,g_1)\to(\M,g_2)$ that preserves the metric $g_1$, the push-forward metric $R_*g_1$ is exactly the metric $g_2$ on $R(\M)$, thus  $R_*g_1=g_2$. This means $R$ is an isometry.

\begin{figure}
    \centering
    \includegraphics[height=0.15\textheight]{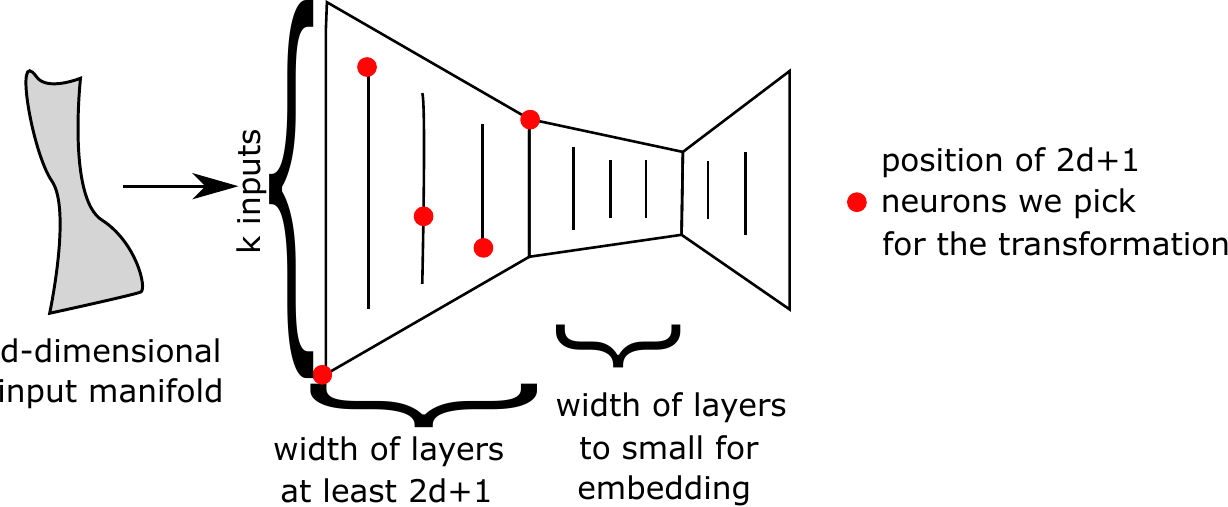}
    \includegraphics[height=0.15\textheight]{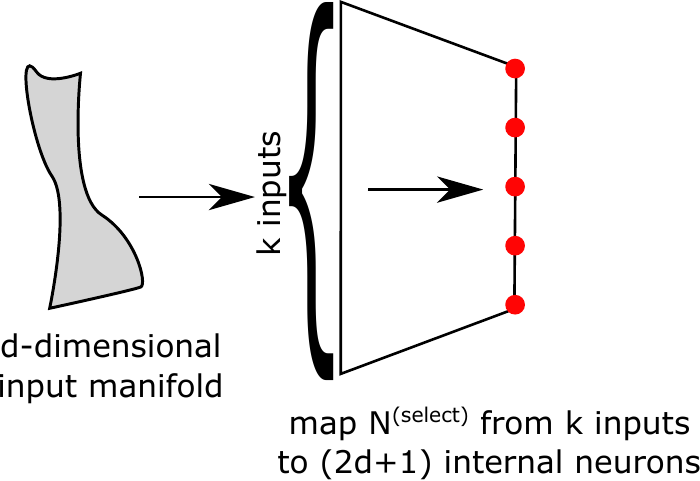}
    \caption{\label{fig:network_layout}
    Left:
    Network layout and selection of $\ell\geq 2d+1$ neurons in the first part of the network (where all layers have enough neurons to embed the $d$-dimensional input manifold $\M$). The activations of the $\ell$ neurons contain enough information to map (a) to the activation of any other neuron, or (b) to any neuron of another network defined on the same manifold.
    Right:
    the map $N^{\text{(select)}}$, from the same $k$ inputs as the network, to the $\ell$ internal neurons (the red dots on the left).
    }
\end{figure}

Let $D=\{x\in\R^k\}\subset\M$ be a finite collection of points on $\M$, sampled from the measure $\mu$; for our first example, this is just the uniform measure on $[-1,1]$.
Two artificial neural networks  $N_1:\R^k\to\R^m$, $N_2:\R^k\to\R^m$ approximate respectively the tasks $f_1$ and $f_2$, such that for all $x\in D$,
$
N_1(x)\approx f_1(x),\ N_2(x)\approx f_2(x).
$
We purposefully omit assumptions on how the two networks generalize for data not contained in $D$, because we want to test these generalization properties later.
We will later also discuss how to transform between networks trained on different input datasets.

We then consider the output of every individual neuron in the two networks as a real valued function on the d-dimensional input manifold.
If every layer has at least $2d+1$ neurons, the topology of the input manifold will, generically, be preserved while it is passed from layer to layer. 
In this case, all neurons everywhere in the network can be considered generic observables of the input manifold, and so picking any $2d+1$ of them as coordinates is enough to create an embedding space for the input manifold.

This means that our transformation function can map from any $2d+1$ neurons in one network to any $2d+1$ neurons in the other network, invertibly, for data on the manifold.

As an illustration of the theoretical concept, consider a line segment $\M=[-1,1]$. If we pick a single polynomial $p(x) =(x-2a)x$ with the parameter $a$ chosen randomly in $(-1,1)$, it cannot be used as an embedding of $\M$, because it folds over at the point $x=a$. However, if we pick three of these random polynomials as three new coordinates for $\M$, then $(p_1(\M), p_2(\M), p_3(\M))$ is an embedding of $\M$ into $\mathbb{R}^3$ almost surely (i.e. with probability one). The fact that this holds for arbitrary smooth manifolds $\M$ and $2d+1$ coordinates follows from the work of~\cite{whitney-1936}, with probabilistic arguments by~\cite{sauer-1991}. 
$2d+1$ generic observations provide a guarantee; if one is lucky, one may be able to create a useful embedding with less (but of course at least $d$!) observations.

In general, if any layer (in particular, the output) of a network has less than $2d+1$ neurons, 
(or if, say, several weights become identically zero so that the output of several neurons in a layer is constant, and no longer a generic function over the input)
then the  topology of the input manifold may not be preserved, and the manifold will be ``folded'' or ``collapsed''. In this case, a transformation to the other network is no longer possible when we only use observations of neurons in or after that layer. We demonstrate this in \S\ref{sec:parabola}.

We will write $N_i^{\text{(select)}}:\mathbb{R}^k\to\mathbb{R}^{\ell}$ for the map of the input of network $i=1,2$ to $\ell\geq 2d+1$ internal neurons that we select for our transformation, and $N_i:\mathbb{R}^k\to\mathbb{R}^m$ for the mapping from the input to the output.
These $l$ neurons must constitute generic observations of the input manifold; they can in principle lie at any internal (some possibly even at the input or output) layer. 
Figure~\ref{fig:network_layout} illustrates that in the network we can pick (red) neurons as the output of our map $N_i^{\text{(select)}}$ only in layers where all previous layers have at least $2d+1$ neurons.

%
%We will take advantage of internal representations of the input data in the networks, and so assume that $N_1$ and $N_2$ have at least one hidden layer each, such that they can be decomposed as 
%$N_1=N_1^{(out)}\circ N_1^{(in)}$ and $N_2=N_2^{(out)}\circ N_2^{(in)},$
%where we assume $N_i^{(in)}:\R^{k}\to\R^{\ell}$, $N_i^{(out)}:\R^{\ell}\to\R^{m}$ for $i=1,2$, and $\ell\geq 2d+1$. This means that the number of internal neurons has to be at least $\ell$.
%
%The decomposition into $N_i^{(in)}$ and $N_i^{(out)}$, and their mapping to and from $\ell$-dimensional Euclidean space, is a basic requirement that the actual neural networks we study have to fulfill.
%We do not require that these networks have exactly one layer with $\ell$ neurons, but that the minimal number of internal neurons per layer is at least $2d+1$---anywhere in the network, across all layers.
% In fact, if we include some of the input coordinates as an input to the transformation function to the other network, then the number of internal neurons per layer can be reduced.
%
%The number of neurons in the hidden layers has to be at least $2d+1$  to preserve the topology of the input manifold, due to the theorem of~\cite{whitney-1936}.
%
%For the construction of the transformation function, we are then able to utilize \textit{any} $2d+1$ internal (or output, or input) coordinates.
%
%The functions $N_i:\mathbb{R}^k\to\mathbb{R}^m$ represent the full mapping of network $i$, from input to output. 
%
The problem of finding transformations between neural networks can now be formulated as follows.
\begin{definition}{Based on output data only.}
Construct an invertible map $T_A:\R^m\to\R^m$ such that $\forall x\in D$,
\begin{equation*}
(T_A\circ N_1)(x)=N_2(x).
\end{equation*}
\end{definition}

\begin{definition}{Based on internal activation data.}
Construct an invertible map $T_B:\R^{\ell}\to\R^{\ell}$, with $\ell\geq 2d+1$, such that $\forall x\in D$, 
\begin{equation*}
\left(T_B\circ N_1^{\text{(select)}}\right)(x)=N_2^{\text{(select)}}(x).
\end{equation*}
\end{definition}

\begin{definition}{Different input data.}
Assume the tasks $f_1$ and $f_2$ are defined on two isometric Riemannian manifolds $(\M,g_1)$  and $(\M,g_2)$ (such that there exists an isometry $R:\M\to\M$ with $R_*g_1=g_2$).
Construct the map $T_B:\R^{\ell}\to\R^{\ell}$, $\ell\geq 2d+1$, such that for $\forall x\in D$,
\begin{equation*}
\left(T_B\circ N_1^{\text{(select)}}\right)(x)=\left(N_2^{\text{(select)}}\circ R\right)(x).
\end{equation*}
\end{definition}
Here, $R_*g_1$ denotes the push-forward of the metric $g_1$ by $R$.

\subsection{Constructing the transformation}\label{sec:constructing transformation}

In this section, we explain how to construct the transformation between the networks in a consistent way.
% 
%  MAHALANOBIS DESCRIPTION %
The crucial step in our approach is to employ spectral representations of the internal activation space of the two networks.
These representations are constructed through manifold learning techniques introduced by \cite{singer-2008,berry-2015}:
Given a metric $g$ and its push-forward metric $S_{\ast}g$ by a diffeomorphism $S: \M \to \M$, the map $S$ can be reconstructed 
up to a linear, orthogonal map. 
The reconstruction can even be done in a data-driven way, 
employing diffusion map embeddings \citep{coifman-2006} and a  so-called ``Mahalanobis distance''~\citep{singer-2008,dsilva-2016}.
%

%%%%%%%%%%%%%%%%%%%%%%%%%%%%%%%%%%%%%%%%%%%%%%%%%%%%%%%%%%%%%%%
% DMAPS
%%%%%%%%%%%%%%%%%%%%%%%%%%%%%%%%%%%%%%%%%%%%%%%%%%%%%%%%%%%%%%%%
The diffusion maps algorithm utilizing the Mahalanobis distance is described in algorithm~\ref{alg:diffusion maps}.
We often only have non-linear observations of points on the manifold $\M$, given by an observation function $S:\M\to\mathbb{R}^m$.
In our case, the tasks $f_1,f_2$ and in particular the neural network maps $N_i^{\text{select}}$ provide such non-linear observations of the input manifold $\M$.
Such maps usually distort the metric on $\M$, so that the original geometry is not preserved. We can still parametrize $\M$ with its original geometry if we employ the following approach: 
In the kernel used by the Mahalanobis Diffusion Map algorithm, the similarity between points includes information about the observation functions $S$ (here, the $N_i^{\text{select}}$) that have to be invertible on their image. For $y_i,y_j\in S(\mathcal{M})\subset\mathbb{R}^m$,
\begin{eqnarray}\label{eq:Mahala_distance}
d^2_M(y_i,y_j)&=&(y_i-y_j)^T\left(J^T(y_i) J(y_i)+J^T(y_j) J(y_j)\right) (y_j-y_i),\\\label{eq:Mahala_kernel}
    k(y_i,y_j)
    &=& \exp\left(-\frac{d_M^2(y_i,y_j)}{2\varepsilon}\right),
\end{eqnarray}
where $J(y)$ is the Jacobian matrix of the inverse transformation $S^{-1}$ at the point $y$.
%This concept was introduced as ``non-linear independent component analysis''~\cite{singer-2008}.
The product $J^T J$, on which the entire procedure hinges, can be approximated through observations of data point neighborhoods: typically in the form of a covariance, obtained, for example, by short bursts of a stochastic dynamical system on $\M$, subsequently observed through $S$~\citep{singer-2009}.
In algorithm~\ref{alg:diffusion maps}, we do not specify how the neighborhoods for the computation of the covariance matrix are obtained. We do so separately for each of the computational experiments in section~\ref{sec:demonstrations} (see also the discussion by~\cite{dietrich-2020}).

The reconstruction of the eigenspace up to an orthogonal map is justified through the following argument from \cite{moosmueller-2020}: using the distance \eqref{eq:Mahala_distance} is effectively using the metric $S_{\ast}g$ on the data, turning $S$ into an isometry. 
Laplace-Beltrami operators of isometric manifolds have the same eigenvalues, and in that case eigenfunctions associated to the same eigenvalue are related by an orthogonal map \citep{berry-2015,rosenberg-1997}.
% \comment{Should this say "eigenfunctions associated to the same eigenvalues"?}
%
%
Therefore, an isometry $S$ between the base manifolds $(\M,g)$ and $(\M,S_*g)$ turns into an orthogonal map in eigenfunction coordinates of the manifolds~\citep{berry-2015}.

% The metric $S_{\ast}g$ can be computed from $g$ by
% $
%     (S_{\ast}g)_y(\xi,\eta) = g_{S^{-1}(y)}(J(y)\xi,J(y)\eta),
% $
% where $\xi,\eta \in T_y\M$ and $J(y)$ denotes the Jacobian of $S^{-1}$ at $y\in \N$. If $\N$ is % embedded in Euclidean space, to use the push-forward metric $S_{\ast}g$ instead of the induced % Euclidean metric on $\N$ for DMAP embeddings, a special kernel
% can be employed~\eqref{eq:Mahala_kernel}. The kernel requires estimations of the Jacobian % matrices of $S$ at every point, as discussed above. The induced metric $S_{\ast}g$ computed via % covariance matrices is also referred to as the \emph{Mahalanobis distance}~\citep{dsilva-2016}, % see equation~\eqref{eq:Mahala_distance}.
%
% Algorithm~\ref{alg:diffusion maps} outlines how a diffusion map can be constructed by using an approximation of that distance.

\begin{algorithm}
\SetAlgoLined
\textbf{Input:}
$n\in\mathbb{N}$: Number of neighborhoods.
$m\in\mathbb{N}$: Dimension of points in the input.
$q\in\mathbb{N}$: Number of points per neigborhood.
$\ell\in\mathbb{N}$: Number of eigenvectors to compute.
$Y$: Set of $n$ neighborhoods $Y_i=\left\lbrace  y_{i}^{(j)}\in {\mathbb{R}^{m}}\right\rbrace$, $i=1,\dots,n$, $j=1,\dots,q$.
\\
\begin{enumerate}
    \item Compute inverse covariance matrices $C_i=\text{cov}(Y_i)^{-1}\in\mathbb{R}^{m\times m}$, $i=1,\dots,n$.
    \item Compute kernel matrix $K\in\mathbb{R}^{n\times n}$ through equation~\eqref{eq:Mahala_kernel}, using the approximation $J(\mathbb{E}[Y_i])^T J(\mathbb{E}[Y_i])\approx C_i$ and kernel bandwidth $\epsilon$ through median of squared Mahalanobis-distances to $k$-th nearest neighbor of all points, where we choose $k=10$.
    \item Normalize kernel matrix to make it invariant to sampling density:
    \begin{enumerate}
        % \item Form the diagonal normalization matrix $P_{ii}=\sum_{j=1}^N K_{ij}$.
        \item Form the kernel matrix $W = P^{-1} K P^{-1}$, with diagonal matrix $P_{ii}=\sum_{j=1}^N K_{ij}$.
        % \item Form the diagonal normalization matrix $Q_{ii} =\sum_{j=1}^N W_{ij}$.
        \item Form the matrix $\hat{T}=Q^{-1/2} W Q^{-1/2}$ with diagonal matrix $Q_{ii} =\sum_{j=1}^N W_{ij}$.
        \item Find the $\ell$ largest eigenvalues $a_p$ associated to non-harmonic eigenvectors $v_p$ of $\hat{T}$. See~\cite{dsilva-2018} for a discussion and method to remove harmonic eigenvectors.
        \item Compute the eigenvalues of $\hat{T}^{1/\epsilon}$ by $\lambda_p=a_p^{1/(2\epsilon)}$.
        \item Compute the eigenvectors $\phi_p$ of the matrix $T = Q^{-1}W$ through $\phi_p=Q^{-1/2}v_p$.
    \end{enumerate}
    \item Define diffusion map $\phi:\mathbb{R}^m\to\mathbb{R}^\ell$ and its inverse through interpolation (here: nearest neighbor interpolation. More advanced techniques such as Nystr\"om extension~\citep{chiavazzo-2014} or neural networks are also possible---this is a supervised learning problem).
\end{enumerate}
\textbf{Output:}
$\phi:\mathbb{R}^m\to\mathbb{R}^\ell$: diffusion map from the input space to $\ell$ eigenvectors.
 \caption{\label{alg:diffusion maps}Diffusion maps algorithm, using the Mahalanobis-metric~\citep{singer-2009} (and the similarity transform from~\cite{berry-2013}).}
\end{algorithm}

Finally, to construct the transformation between neural networks using the diffusion map approach, we proceed as described in algorithm~\ref{alg:transformation}. Essentially, we generate embeddings of the input manifolds of the two networks using their internal neurons, and then employ algorithm~\ref{alg:diffusion maps} to construct a consistent representation for each of these embeddings. Since these representations are invariant up to an orthogonal map, we can construct our transformation after estimating this final map, which only requires a small number of ``common measurements'' (it is easier to parametrize linear maps than general nonlinear ones!).
%
% The next section demonstrates our approach to construct the transformation for several examples.

\begin{algorithm}
\SetAlgoLined
\textbf{Input:}
$\ell\geq 2d+1$: Number of neurons to use from both networks $N_1, N_2$.
$\ell$ is also the number of eigenvectors we use for the spectral embedding.
$S_1,S_2:\mathcal{M}\to\mathbb{R}^{\ell\times q}$: Sampling procedures of neighborhoods, consisting of $q\in\mathbb{N}$ points each, on the input manifolds.
$n_1,n_2\in\mathbb{N}$: Numbers of neighborhoods to sample using $S_1,S_2$.
\\
\begin{enumerate}
\item Sample sets of neighborhoods $\mathcal{X}_1=\left\lbrace S_1(i)\right\rbrace_{i=1}^{n_1}$, and $\mathcal{X}_2=\left\lbrace S_2(i)\right\rbrace_{i=1}^{n_2}$.
\item Evaluate the networks $N_1,N_2$ separately on all points in their sets of $\mathcal{X}_1,\mathcal{X}_2$ and record the output of the $m$ selected neurons in sets of neighborhoods  $\mathcal{Y}_1,\mathcal{Y}_2$.
\item Use the neighborhoods in $\mathcal{Y}_1,\mathcal{Y}_2$ for algorithm (\ref{alg:diffusion maps}) to construct invertible maps $\netOneDMAP,\netTwoDMAP$ from data of the $m$ neurons of $N_1,N_2$ to the embedding into $\ell$ eigenvectors (see algorithm~\ref{alg:diffusion maps}).
\item There is only an orthonormal transformation $O\in SO(\ell)$ missing between the two sets of eigenvectors. Here, we estimate it with a few known corresponding points. Alternatives include the use of an iterative closest point algorithm~\citep{rusinkiewicz-2001} or Procrustes alignment methods, e.g.~\citep{kabsch-1976}.
\item Construct the map $T$ through 
$
    y_2=T(y_1)=\netTwoDMAP^{-1} \circ O \circ \netOneDMAP(y_1).
$
\end{enumerate}
\textbf{Output:}
{Invertible map $T:\mathbb{R}^{\ell}\to\mathbb{R}^{\ell}$ from $\ell$ neurons in network $N_1$ to $\ell$ neurons in $N_2$.}
 \caption{\label{alg:transformation}Algorithm to compute the transformation between two neural networks.}
\end{algorithm}

%%%%%%%%%%%%%%%%%%%%%%%%%%%%%%%%%%%%%%%%%%%%%%%%%%%%

\section{Demonstrating transformations between neural networks}\label{sec:demonstrations}
%
% \begin{figure}[ht!]
%     \centering
%     \includegraphics[width=.75\textwidth]{figures/network_mapping}
%     \caption{If we can find a way to make the connection, we could discuss this "mapping between networks"/"network debriefing" technique here.}
%     \label{fig:network_mapping}
% \end{figure}

We now demonstrate the transformation concept in three separate examples. %, from simple to complex.
Section~\ref{sec:parabola} illustrates a simple transformation between two networks with the same one-dimensional input, trained on the same task.
Section~\ref{sec:differentInputs} then broadens the scope to networks trained on different one-dimensional inputs.
Section~\ref{sec:vector fields} shows how to construct a transformation using observations of internal activations of two networks defined on two-dimensional spaces.
Section~\ref{sec:images} describes our most intricate example, with images as input for two deep auto-encoder networks with several convolutional layers. We show how to map between internal activations of these networks and how to reconstruct the output of one network from the input of the other.
Table~\ref{tab:experiments data} contains metrics and parameters for the computational experiments.

\newcommand{\hiddenNOne}{1}
\newcommand{\hiddenNOneWord}{one}
\newcommand{\hiddenNTwo}{8}
\newcommand{\hiddenNTwoWord}{eight}
\newcommand{\scin}[2]{$#1\cdot10^{#2}$}
\newcommand{\imageTrainLossA}{\scin{1.2}{-1}}
\newcommand{\imageTrainLossB}{\scin{1.1}{-1}}
\newcommand{\imageValLossA}{\scin{2.6}{-1}}
\newcommand{\imageValLossB}{\scin{1.7}{-1}}
\newcommand{\hang}{\hangpara{.03\textwidth}{1}}
\begin{table}[h]
  \caption{Data and parameters of computational experiments. If not otherwise specified, we use the same parameters for both networks in a column; otherwise, different network hyperparameters are given on different lines. Activation functions are $\tanh$ except for \S\ref{sec:images}, as described in the text.}
  \label{tab:experiments data}
  \centering
  \begin{tabular}{p{0.17\textwidth} | p{0.16\textwidth} p{0.14\textwidth} p{0.20\textwidth} p{0.17\textwidth}}
    \toprule
    Experiment     & \S~\ref{sec:parabola}  & \S~\ref{sec:differentInputs} 
    & \S~\ref{sec:vector fields} & \S~\ref{sec:images} \\
    \midrule
    \begin{hang}\# Training\end{hang}
        & 10,240                                               % Vanilla parabolas
        & 900                                                 % Different inputs                 
    % \todo{really only 12 points? wow--well it is a parabola after all.}\comment{Really. \texttt{
       %  https://www.dropbox.com/s/v6t8bv7zfeoc1kw/matching\%20with\%20different\%20input\%20domains.ipynb?dl=0
       %  }
       %  }                     
        & 3,600                                                % Vector fields
        & 1,943\newline\tiny (at 64$\times$64 resolution)      % images
    \\ \# Validation
        & 512 
        & 100
        & 400
        & 216
    \\ Training Error
        & \scin{5.7}{-6} \newline \scin{2.0}{-5}
        & \scin{1.48}{-5} \newline \scin{8.60}{-3}
        & \scin{7.3}{-5} \newline \scin{2.5}{-6}
        & \imageTrainLossA{} \newline \imageTrainLossB{}
    \\ Validation Error
        & \scin{1.4}{-1} \newline \scin{1.0}{-1} 
        & \scin{1.58}{-5} \newline \scin{5.66}{-3}
        & \scin{4.6}{-5} \newline \scin{2.4}{-6}
        & \imageValLossA{} \newline \imageTrainLossB{}
    \\ Architecture
        & 1-\hiddenNOne-1, 1-8-8-\hiddenNTwo-1
         & 1-3-1
        & 2-5-5-5-5-2
        & {CNN (see text)} 
    \\ \midrule
    % \hang
    $q$: Nbhd. size
        & 1,024
         &  \num{11}%$=2\cdot d+1$
        & \small (analytic covariance)
        &  431
    \\ 
    % \hang
    $n$: \# Nbhds.
        & 512
         & \num{990}%$=n-d-d$
        & 100,000 
        & 649 \\
     \hang
     $\epsilon$: Kernel \newline bandwidth
        & \scin{1.33}{1}\newline \scin{5.54}{1}
         & 5
        & \scin{1.21}{-3}\newline\scin{8.06}{-4} 
        & \scin{9.68}{0}\newline\scin{1.33}{1}
    \\ \bottomrule
  \end{tabular}
\end{table}

\subsection{Transformations between output layers of simple networks}\label{sec:parabola}

In this first example, we demonstrate the idea of transformations between neural networks in a very simple caricature: 
Two tasks $f_1,f_2:\mathcal{M}\to\mathbb{R}$ with $\mathcal{M}=[-1,1]\subset\mathbb{R}$, are approximated by two neural networks $N_1,N_2$.
% 
% In this setting, we can even visualize the full, three-dimensional activation space we use, as well as the manifolds we are mapping between (one-dimensional, connected subsets of Euclidean space).
%
% We first show how to transform between output layers of two networks in \S\ref{sec:transformedOutputs}, and then in~\S\ref{sec:differentInputs} construct a transformation if the input manifolds and the tasks are different.
% 
% \subsubsection{Transformation between output layers}
% \label{sec:transformedOutputs}
We choose the tasks to be identical, $f_1=f_2=f$, and defined by $f(x)=-x^2.$ 
The training data set for the two networks is also the same, $D_1=D_2=D\subset [-1,0].$
We follow algorithm~\ref{alg:transformation} to construct the map $T:\mathbb{R}\to\mathbb{R}$ between the one-dimensional output of both networks. The local neighborhoods $Y_i$ of points in the training domain $D_1$ are generated through $\delta$-balls around each point, with $\delta=0.05$.

\begin{figure}
% /home/tsbertalan/Dropbox/Projects/Representations, Debriefing, and Mapping/Compare nonmonotonic networks.ipynb
    \centering
    \begin{tabular}{cc}
    \includegraphics[width=.35\textwidth]{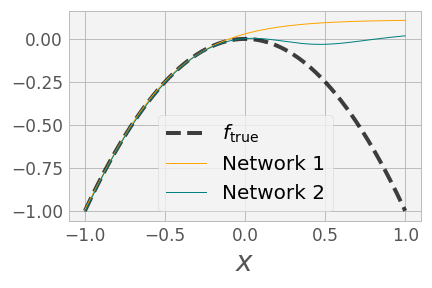}&
    \raisebox{.3cm}{\includegraphics[width=.4\textwidth]{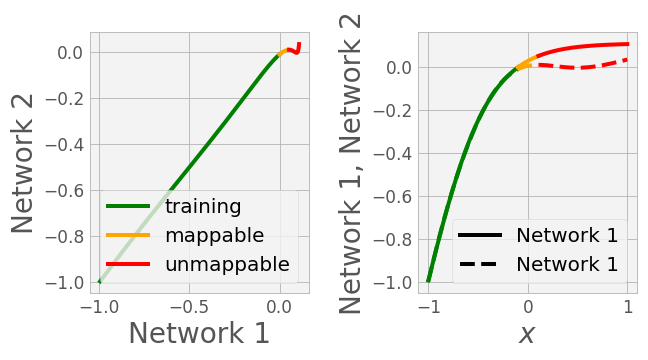}}\\
    (a)&(b)
    \end{tabular}
    \caption{(a) Two networks $\Net_1$ and $\Net_2$ fitted to the left half of a parabola (dashed). $\Net_1$ has one hidden $\tanh$ neuron, and so is strictly monotonic w.r.t. the input $x$, and $\Net_2$ has 24 neurons in three hidden layers.
(b) Output of the two networks plotted against each other. They can be mapped invertibly within the training data (green) as well as in a small neighborhood beyond (yellow).
}
    \label{fig:parabola1d}
\end{figure}

\begin{figure}
% also /home/tsbertalan/Dropbox/Projects/Representations, Debriefing, and Mapping/Compare nonmonotonic networks.ipynb
    \centering
    \begin{tabular}{ccc}
    \includegraphics[width=.3\textwidth]{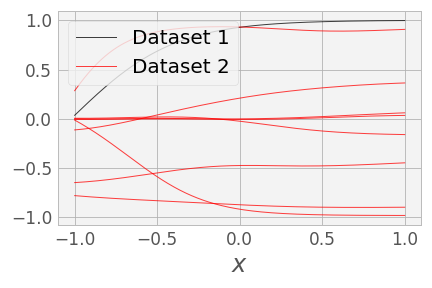}&
    \includegraphics[width=.35\textwidth]{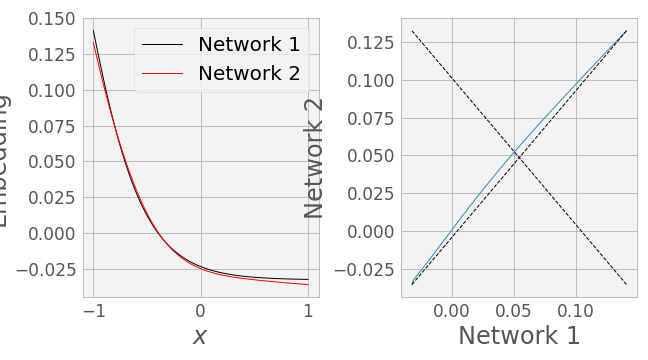}
    &
    \includegraphics[width=.3\textwidth]{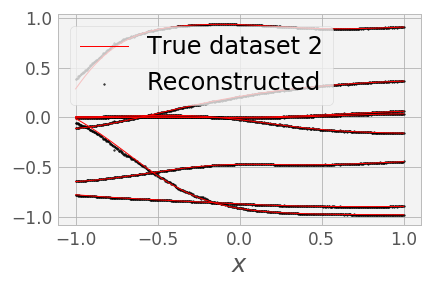}
    \\
    (a)&(b)&(c)
    \end{tabular}
    \caption{
    (a) Activations of neurons in the hidden layers of the two networks, evaluated beyond the training region. Dataset 1 contains the activations of the single hidden neuron in $\Net_1$; dataset 2 contains the activations of the last hidden layer of $\Net_2$. 
    (b) Separately embedding the two datasets with diffusion maps (and Mahalanobis scaling) results in coordinates that are related by an orthogonal transformation (in one dimension, multiplication by $\pm 1$).
    (c)
    Our selected activations from the second network, reconstructed as functions of the selected activations from the first network.
}
    \label{fig:parabola1d extended input}
\end{figure}

Figure~\ref{fig:parabola1d}(a) shows the task $f$ as well as the evaluations of the two networks on the training data in $[-1,0]$ as well as their extrapolation to all of $\mathcal{M}$. The networks approximate $f$ well on the training data, but extrapolate differently beyond it.
This extrapolation issue is further illustrated in Fig.~\ref{fig:parabola1d}(b), where we can see that it is impossible to map from network $N_2$ back to network $N_1$ over all of $\M$, because the output of $N_2$ is not an invertible function of the output of $N_1$ over all of $\M$.
However, if we use as observables the \hiddenNOneWord{} or \hiddenNTwoWord{} neuron(s) in the final hidden layers of the two networks (see Fig.~\ref{fig:parabola1d extended input}(a), and architectures in Table~\ref{tab:experiments data}), we can easily construct the transformation over all of $\M$ using algorithm~\ref{alg:transformation} (for the result see Fig.~\ref{fig:parabola1d extended input}(b)).
Fig.~\ref{fig:parabola1d extended input}(c) shows the reconstruction of the selected neuron activations of Network 2 based on our transformation and our Network 1 activation observations. It is worth noting that {\em any} other activation of Network 2 neurons can be also approximated through, e.g. Geometric Harmonics as functions of the intrinsic representation; other types of approximation of these functions (e.g. neural networks or Gaussian Processes, or even just nearest neighbor interpolation) are also possible. 
For the purposes of Fig.~\ref{fig:parabola1d extended input}(c), we compute $\netTwoDMAP^{-1}$ through univariate, linear interpolation.
Note that for Network 1 we can ``get by" with a single observation and not the ``guaranteed" $2 \times 1 + 1 =3$ observations: the architecture and the  activation function are so simple that this single observation is one-to-one with the single input (does not ``fold" over the input). 

\subsection{Transformation between internal neurons of networks with different inputs}\label{sec:differentInputs}
%
% \begin{figure}[ht!]
%     \centering
%     \includegraphics[width=.75\textwidth]{figures/network_mapping}
%     \caption{If we can find a way to make the connection, we could discuss this "mapping between networks"/"network debriefing" technique here.}
%     \label{fig:network_mapping}
% \end{figure}

Here, we demonstrate the transformation concept through a simple mapping between two networks trained on the same task, but with {\em different}, albeit still one-dimensional inputs.
This is similar to experiment \S{4.3}, but in a much lower-dimensional ambient space, where we can easily visualize all embeddings.
Table~\ref{tab:experiments data} contains metrics and parameters for this example.

%  \begin{figure}[ht!]
%  \centering
%  \begin{tabular}{cc}
%  \includegraphics[width=.45\textwidth]{figures/untransformed_coordinates}&
%  \includegraphics[width=.45\textwidth]{figures/transformed_coordinate_system}\\
%  (a)&(b)
%  \end{tabular}
%  \caption{\label{fig:input transform 1}\todo{Remove this figure?}(a) Functions % $f_1$ and $f_2$ over the input axis $x_1$. b) the graphs of $f_1$ and $f_2$ % transformed by rotation and translation, from the original coordinates % $(x_1,y_2)$ into a new coordinate system $(x_2,y_2)$.}
%  \end{figure}

 %\subsubsection{Transformation between internal neurons of networks with different inputs}
 %\label{sec:differentInputs}
 
\begin{figure}
 %\begin{tabular}{cc}
  \centering
  \includegraphics[width=.35\textwidth]{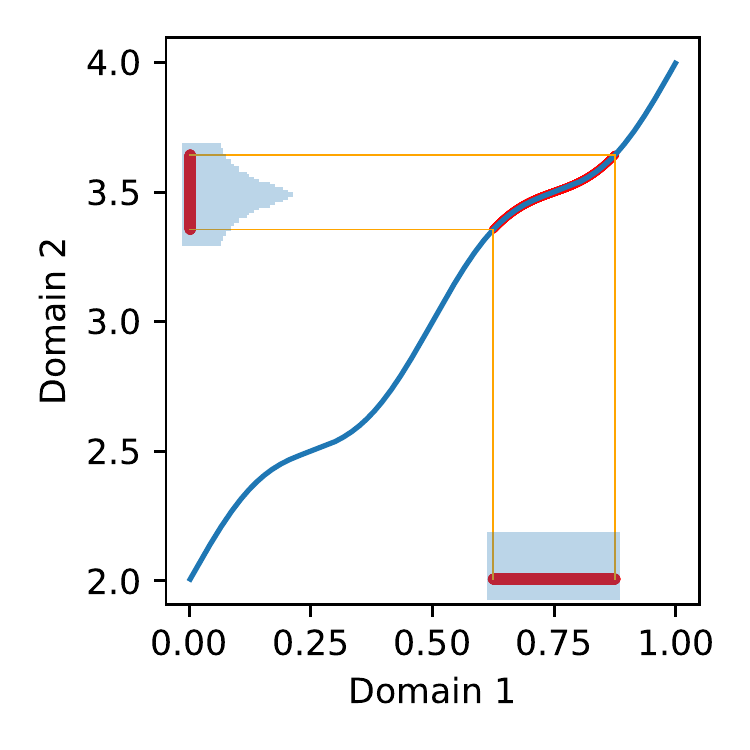}
 \caption{\label{fig:mahalanobis_concept}Uniform density in a local neighborhood on domain 1 is being transformed by the function $S$ to non-uniform density on domain 2. Using the covariance of the neighborhoods in domain 2, we can estimate the derivative (in higher dimensions, Jacobian matrix) of the transformation function between the domains.}
 %\end{tabular}
\end{figure}

%  This example illustrates how to transform between two neural networks trained on different inputs and for different portions of a task.
 %
 Define tasks $f_1,f_2$ identical to $f:\mathbb{R}\to\mathbb{R}$ with $f(x)=-(x-0.25)^2-1$.
 We use training data $D_1\in [0,1]$ and $D_2\in [2,4]$, which are related by the nonlinear transformation $S(x)=2+2x+\sin(4\pi x)/(3\pi)$ (see figure~\ref{fig:mahalanobis_concept}).
 This transformation {\em is not an isometry}, but we assume that we have access to neighborhoods created by sampling in domain 1, and then transforming to domain 2 through $S$ (see figure~\ref{fig:mahalanobis_concept}).
 
\begin{figure}
 %\begin{tabular}{cc}
  \centering
  \includegraphics[width=.4\textwidth]{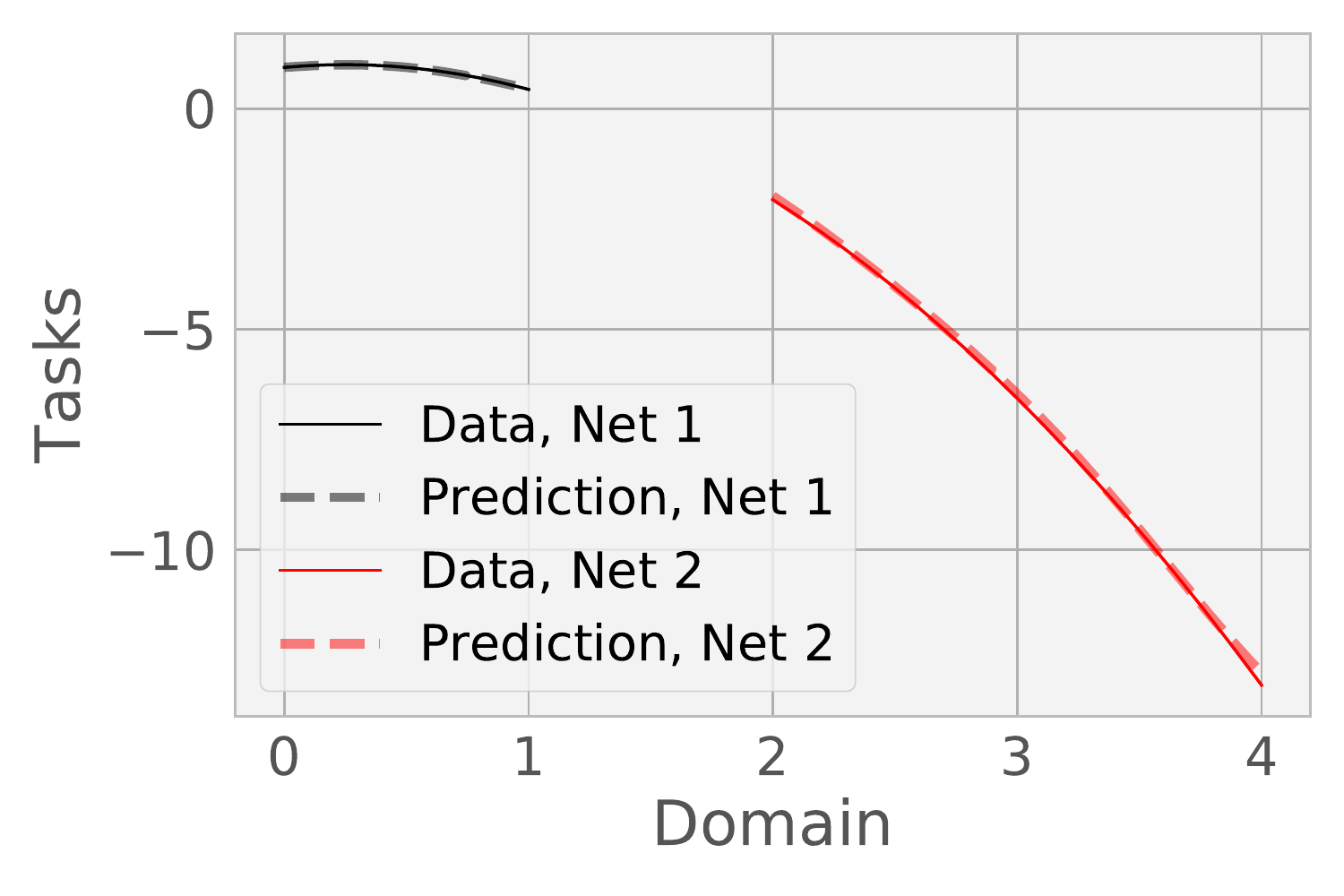}
 \caption{\label{fig:setting}Setting of the computational experiment, showing tasks $f_1,f_2=f$ over the training data $D_1\in[0,1]$ and $S(D_1)=D_2\in[2,4]$.}
 %\end{tabular}
\end{figure}

 The neural networks $\Net_1,\Net_2:\R\to\R$ approximate the task well on their training data (see figure~\ref{fig:setting}).
 To construct the transformation between their internal neurons, we sample small neighborhoods of 15 points for each data point on the domains of the two networks (inside, but also far outside the training data set, on $[0,4]$). The neighborhoods on the domain of the second network are created by transforming points on the domain of the first through $S$.
 It is important to note that the data for our construction of the transformation between the two networks do not need to be evaluated at exactly ``corresponding points", obtained from each other through $S$;  correctly sampled neighborhoods are enough.
 That means we do not even need access to the map $S$, in principle, as long as we are given correct neighborhood information around each point by any other means, e.g. through a consistent sampling procedure~
 \cite{singer-2009,dietrich-2020,moosmueller-2020}.
 We then compute covariance matrices in the space of some triplets of internal neurons of the networks, after evaluating the networks on each point within every data point neighborhood. We do not need to store the output values of the networks, just values of activations of  three neurons anywhere across the network architecture is enough ($2d+1=3$ here, since the input manifold is one-dimensional).
 The covariance matrices are then used to construct the Mahalanobis distance between the data points. As we use three internal neurons for each network, the matrices are $3\times 3$ and all have rank 1.
 As we demonstrate in \S{4.2}, if the inputs are isometric, the covariance matrices with respect to the inputs do not need to be generated in such a way for the approach to work: we can directly compute $J(y)^TJ(y)$ by automatically differentiating through the network. 
% Here, since the inputs are not isometric, we still approximate the Jacobian product to illustrate this separate approach.
 
\begin{figure}
 % /home/tsbertalan/Dropbox/Projects/Representations, Debriefing, and % Mapping/matching with different input domains.ipynb
 \centering
 \begin{tabular}{ p{0.4\textwidth} p{0.25\textwidth} p{0.25\textwidth}}
 \includegraphics[height=.15\textheight]{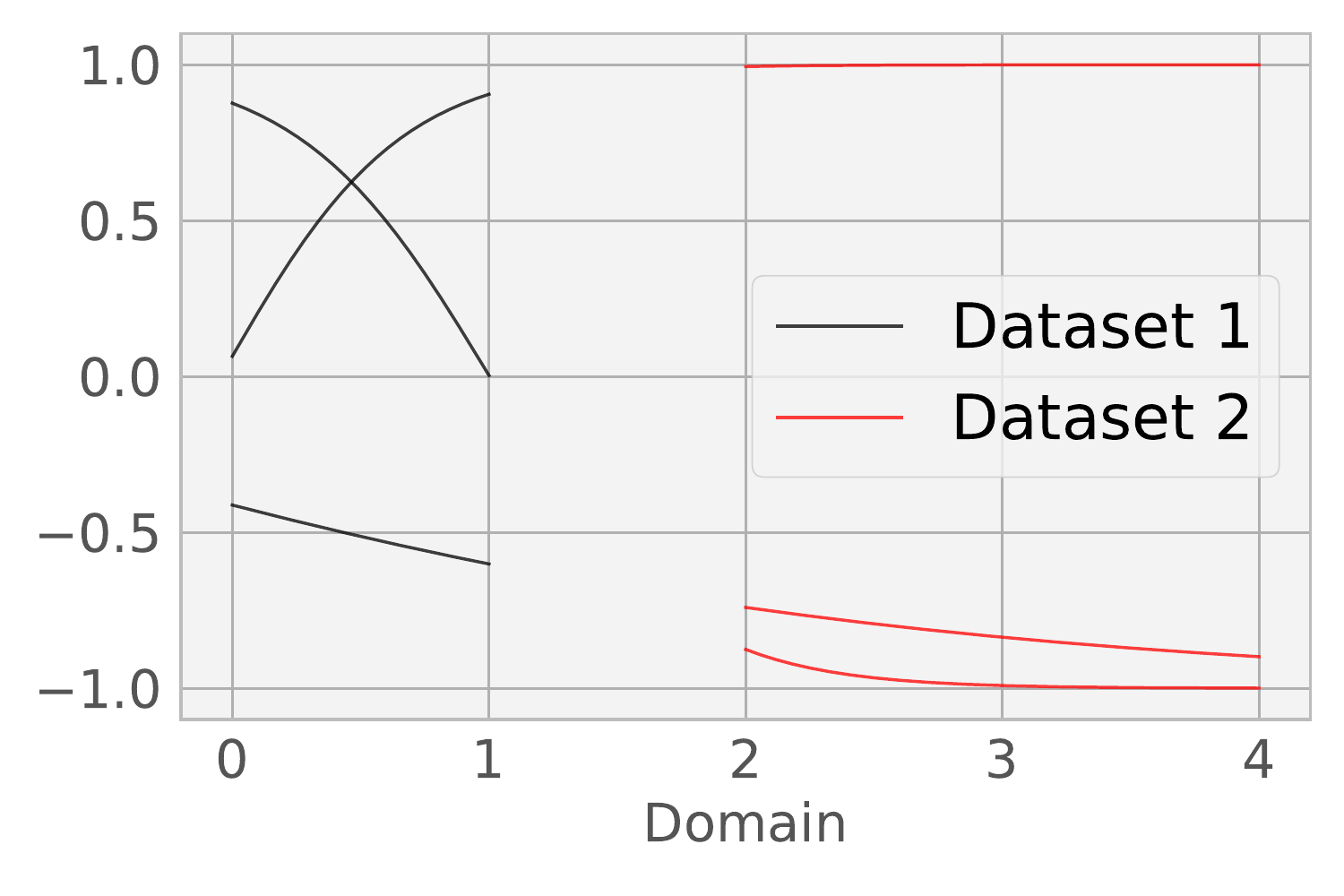}
 &
 \includegraphics[height=.15\textheight]{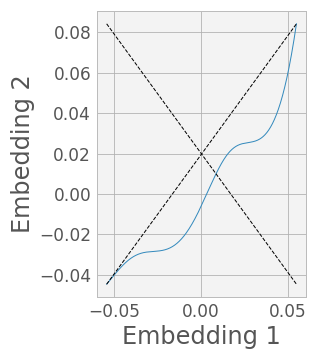}
 & 
 \includegraphics[height=.15\textheight]{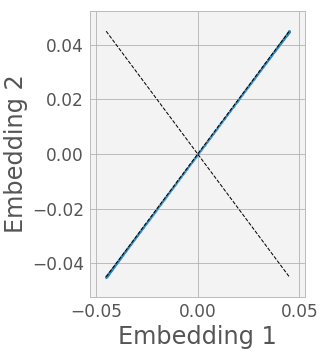}\\
 \centering{(a)}&\centering{(b)}&\centering{(c)}
% \\
%  \begin{tabular}{ p{0.24\textwidth} p{0.24\textwidth} }
%  \multicolumn{2}{c}{\includegraphics[width=.5\textwidth]{figures/v% anilla_DMAP.png}} \\
%  \centering(b)&
%  \centering(c)
%  \end{tabular}
%  & 
%  \begin{tabular}{ p{0.24\textwidth} p{0.24\textwidth}}
%  \multicolumn{2}{c}{\includegraphics[width=.5\textwidth]{figures/M% DMAP.png}} \\
% % \centering(d)&
% %  \centering(e)
%  \end{tabular}
 \end{tabular}
 \caption{\label{fig:activation_space} 
 Panel (a): Selected activations from networks 1 and 2, used as inputs for Mahalanobis diffusion maps.
 %
%  Panel (b): Euclidean diffusion map embeddings of the activation space of the two networks, evaluated over domain 1.
 Panel (b): Using Euclidean distances in the Diffusion Map kernel only recovers embeddings  up to a diffeomorphism.
%  Panel(d): Mahalanobis diffusion map embeddings of the activation space of the two networks, plotted separately over domain 1 and domain 2 $=S($ domain 1$)$ respectively. Note that the density is uniform in domain 1, but not in domain 2.
 %
 Panel (c): Using the Mahalanobis distance recovers almost the same embedding for both networks---up to a sign in one dimension, corresponding to an orthogonal transformation in higher dimensions.}
 \end{figure}

 An embedding of the activations data over the domain $[0,4]$ for the two networks, using diffusion maps and the Mahalanobis distance, results in an embedding of the line segment, related only by an orthogonal map (here, a flip through multiplication by $\pm 1$, see figure~\ref{fig:activation_space}(c)).
 If we do not use the Mahalanobis distance, the embeddings of the input domain into the activation spaces of the two networks generically induce different metrics; then, even though the diffusion map algorithm will result in parametrizations of the two curves, they cannot easily be mapped to each other (figure \ref{fig:activation_space}(b)): they  are related by an arbitrary diffeomorphism, which, in general, a few corresponding points are not enough to approximate.

\subsection{Toward more complicated networks: mapping between vector fields}\label{sec:vector fields}
\begin{figure}
% /home/tsbertalan/Dropbox/Projects/Representations, Debriefing, and Mapping/Mapping between two functions of 2 inputs.ipynb
\centering
\begin{tabular}{ccc}
\raisebox{1.6cm}{(a)}&\includegraphics[height=.17\textheight]{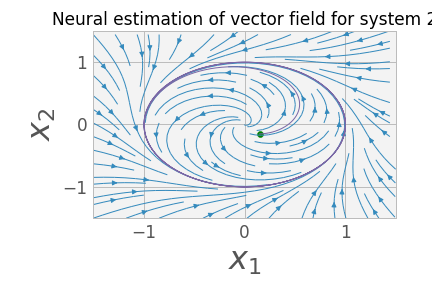}& 
\includegraphics[height=.17\textheight]{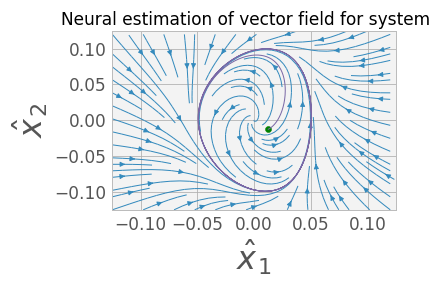}
\\
\raisebox{1.6cm}{(b)}&\multicolumn{2}{c}{\includegraphics[height=.17\textheight]{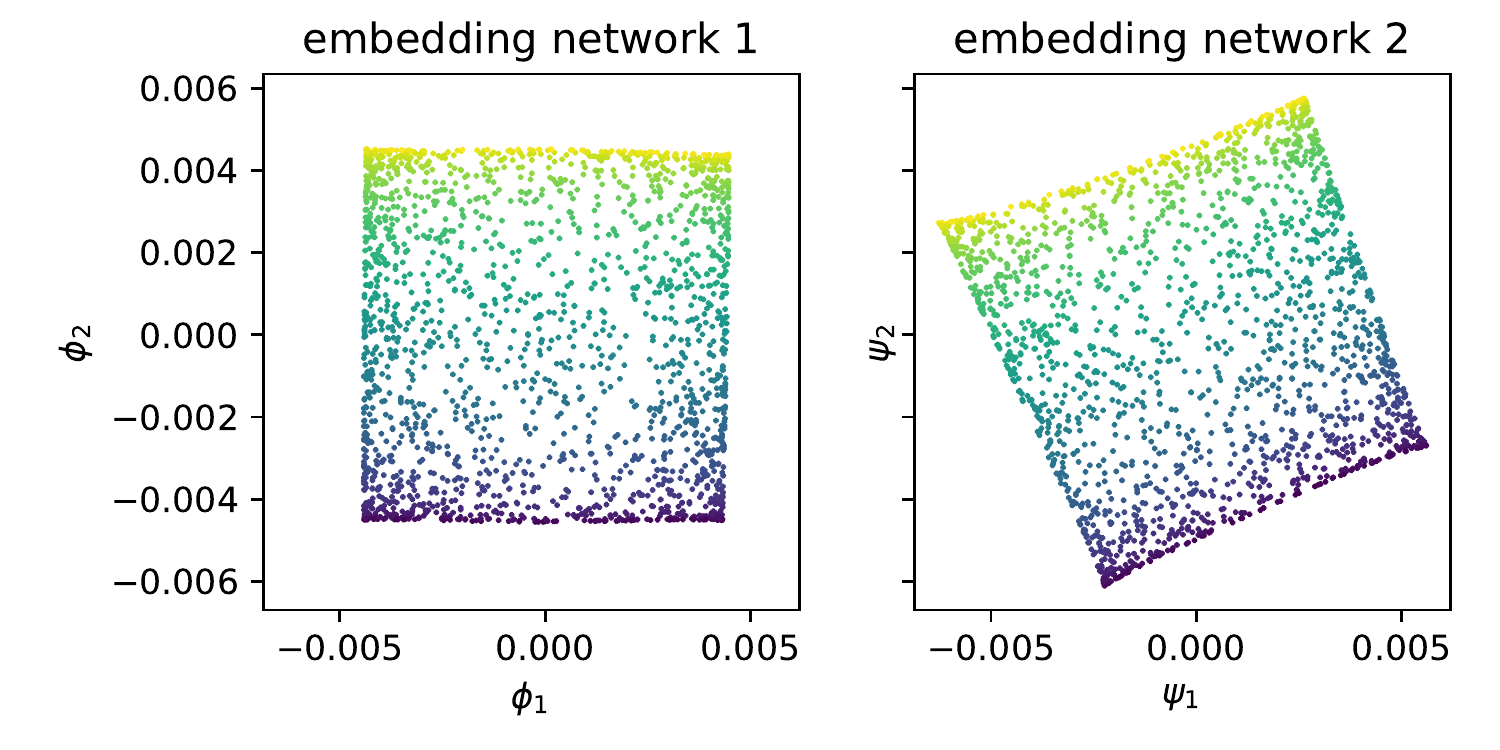}}
\end{tabular}
\caption{
\label{fig:lc2d_vectorfield_data_and_reconstruction}
    % \label{fig:lc2d}
    (a): 
    % data sets and
    Neural network approximations of the vector fields.
    % \todo{Condense Figure \ref{fig:lc2d_vectorfield_data_and_reconstruction} to two panels.}
    % 
    (b): % \label{fig:lc2d_mdmap_embeddings}
    Embedding of the 5-dimensional activation data from the two networks from section~\ref{sec:vector fields}. The covariance matrices were obtained through automatic differentiation. Each panel shows 2,000 points out of the 100,000 used for the embedding, which is consistent for both data sets, and (when ignoring some problems at the edges of the square on the right panel) differs only up to an orthogonal transformation (here, a rotation). The color corresponds to the $x_1$ coordinate (right panel: $\hat x_1$)  in the original space.
}
\end{figure}

We now demonstrate that we can also transform between networks that were trained to approximate vector fields on two-dimensional Euclidean space.  We denote the input coordinates for network 1 by $(x_1,x_2)$ and for network 2 by $(\hat{x}_1,\hat{x}_2)$. Figure~\ref{fig:lc2d_vectorfield_data_and_reconstruction}(a) shows the vector fields\footnote{The vector fields used for the tasks of this experiment are defined via
\begin{equation}
% https://www.wolframcloud.com/env/bertalan0/Transformed%20Dynamical%20Systems.nb
\label{eqn:circle}
\begin{array}{rcl}
    \frac{\der}{\der t}\left(\begin{matrix}x_1\\x_2\end{matrix}\right)&
    =&\left(\begin{matrix}-x_2+x_1(1-x_1^2-x_2^2)\\x_1+x_2(1-x_1^2-x_2^2)\end{matrix}\right),
    \\
    \frac{\der}{\der t}\left(\begin{matrix}\hat{x}_1\\\hat{x}_2\end{matrix}\right)&
    =&J(s^{-1}(\hat{x}_1,\hat{x}_2))\cdot \frac{\der}{\der t}\left(\begin{matrix}s^{-1}(\hat{x}_1,\hat{x}_2)\end{matrix}\right)=J(x_1,x_2) \frac{\der}{\der t}\left(\begin{matrix}x_1\\x_2\end{matrix}\right),
\end{array}
%
%\begin{array}{cc}
%    \begin{array}{rcl}
%         r = \sqrt{{x_1} ^ 2 + {x_2} ^ 2}
%         &&
%         \theta =\mathrm{arctan2}({x_2}, {x_1})
%         \\
%         \frac{\der r}{\der t} = r - r^3
%         &&
%         \frac{\der\theta}{\der t} = 1
%         \\
%         J &=&
%         \frac{\der[{x_1},{x_2}]}{\der[r\theta]}
%        %  J = d([x,y])/d([r,th] = [dxdr, dxdth], [dydr, dydth]
%        \\
%        \frac{\der[{x_1},{x_2}]}{\der t} &=& J \cdot \frac{\der %[r,\theta]}{\der t}
%        %  d[x,y])/dt = J \cdot d([r,th])/dt
%    \end{array}
%     & 
%    \begin{array}{rcl}
%         \hat{x}_1 &=& \frac{{x_1}}{\alpha} + \left( %\frac{{x_1}^2}{\alpha^2} + \frac{{x_2}}{\beta} \right)
%         \\
%         \hat{x}_2 &=& \frac{{x_1}^2}{\alpha^2}+\frac{{x_2}}{\beta }
%    \end{array}     
% \end{array}
\end{equation}
where $J(x_1,x_2)$ is the Jacobian matrix of the map $(\hat{x}_1,\hat{x}_2)=s(x_1,x_2)=\left(\frac{{x_1}}{\alpha} + \left( \frac{{x_1}^2}{\alpha^2} + \frac{{x_2}}{\beta} \right), \frac{{x_1}^2}{\alpha^2}+\frac{{x_2}}{\beta }\right)$, for $\alpha=20$ and $\beta=10$.} we use as tasks to train the neural networks.
%
% It is important to stress that in this experiment, the tasks as well as the output layers of the neural networks are not relevant.
We want to demonstrate that we can construct a transformation $T$ between activations of {\em internal} neurons in case the input manifold has a dimension larger than one (here: two).
The input manifold for the two networks is chosen to be the same: a square $[-1.5,1.5]^2$ in the two-dimensional Euclidean plane, centered on the origin.
The two networks are trained to map points in this two-dimensional domain to the time derivatives at those points, i.e. to coordinates of vectors (also two-dimensional). %
% Both networks have four hidden $\tanh$ layers of width 5. I do a width-2-bottleneck autoencoder of the final 5-neuron hidden layer for each.
%
We explicitly compute the Jacobian product in the Mahalanobis distance \eqref{eq:Mahala_distance}, through automatic differentiation through the networks up to the (five) internal neurons of each.
Densely sampling the input of both networks with 100,000 points leads to as many covariance matrices of shape $5\times 5$, all of rank two (because the input space is two-dimensional).
We sampled that many points to demonstrate that the approach scales well: the construction of the two spectral representations (using algorithm~\ref{alg:diffusion maps}, see figure~\ref{fig:lc2d_vectorfield_data_and_reconstruction}(b)) only took 20 minutes on a single core (Intel Core i9-9900 CPU 3.10GHz).

\subsection{Transformations between deep convolutional networks with high-dimensional input}\label{sec:images}

\newcommand{\lowAngle}{\pi}
\newcommand{\hiAngle}{0.6\pi}

\begin{figure}
    \centering
    \includegraphics[width=1\textwidth]{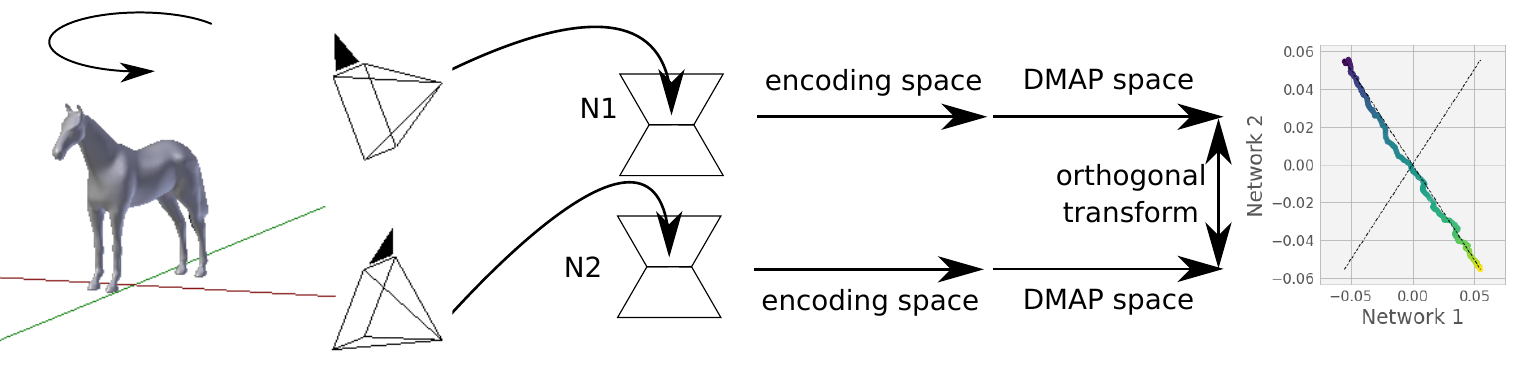}
    \caption{
    Setting for our example of high-dimensional input space. Two cameras observe a rotating horse. We encode the images using two deep autoencoder networks $N_1$ and $N_2$, encoding the images in six dimensions. diffusion maps with the Mahalanobis distance applied to the data in this space yields a consistent embedding, invariant up to an orthogonal transformation (here, a sign flip).
    }
    \label{fig:horse setting}
\end{figure}

In order to demonstrate our transformation between networks on a more realistic example, we generate two datasets of images $x_i$ (high-dimensional representations) with a low-dimensional intrinsic space.
Namely, we rotate a 3D model horse, and render images from two camera angles (see figure~\ref{fig:horse setting}).
For training, we additionally create augmented data by adding $z=|\hat z|$ to each pixel uniformly at random, where $\hat z\sim\mathcal N(0, 1/10)$, pixel values are chosen in $[0,1]$, and we re-threshold pixels to this range after augmentation.
We divide the range $[0,2\pi]$ of object rotation into 3,600 bins, but only train on images in $\theta\in[\lowAngle, \hiAngle]$.

The functions $f_1$ and $f_2$ in this example are denoising tasks for images obtained from the two cameras. To approximate them, we train two convolutional autoencoders to reconstruct the input images,
with the reconstruction target being the noiseless original images
when noise-augmented images are the input.
We minimize a pixelwise binary cross-entropy loss,
to which we add the term $10^{-5}\cdot
\left[
(l-l^*)^2
+
(h-h^*)^2
\right]$,
where $l$ and $h$ are, respectively, the batchwise minimum and maximum of the bottleneck layer activations,
and
$l^*=\lowAngle$ and $h^*=\hiAngle$
are our desired values for this range.
This pinning term does not affect any of our other results, but improves legibility of the plots.

Our encoders have interleaved two-dimensional convolutional layers 
(
    filter sizes $9, 7, 5, 3, 3$,
    channel counts $8, 32, 64, 16$,
    stride of 1,
    and
    ``same'' padding
)
and width-2 max-pooling layers in the encoding portion,
followed by two width-6 dense layers
to output our encodings $y_i$
The decoders have
interleaved convolutional layers 
(with
reversed filter sizes and channel counts to produce a pyramid of same-shape tensors)
and unlearned nearest-neighbor resizing operations in the decoding portion
to
output tensors of the same shapes as the input.
All activations are ReLU
except for the final layer, which is sigmoid.
We use AdaM with Nesterov momentum with a batch size of 128, for 220 epochs, with a learning rate of $10^{-4}$.
Training generally takes less than five minutes per autoencoder on an Nvidia RTX 2080 ti GPU,
with a training loss of \imageTrainLossA{} and validation loss of \imageValLossA{}.

We did not do any systematic hyperparameter search, and there are obvious improvements that we did not attempt to achieve better performance on the autoencoder tasks, as the network$\rightarrow$network transformation was the actual project of interest.

In order to compute the inverse covariances $C_i$ for the Mahalanobis distances described in algorithm~\ref{alg:diffusion maps},
for each (high-dimensional) image $x_i(\theta_i)$
we evaluate the (low-dimensional) encodings $y_j$ for the pre-sampled images $x_j(\theta_j),\,j=i-d,\ldots,i,\ldots,i+d$ to the left and to the right in $\theta$, where $i$ is taken s.t. $j$ never indexes outside the training data 
($d=(q-1)/2$ and $q$ is as given in Table~\ref{tab:experiments data}).
We then evaluate the covariance of this cloud of low-dimensional encodings,
and use the pseudoinverse of this as $C_i$.

\begin{figure}
% /home/tsbertalan/Dropbox/Projects/Representations, Debriefing, and Mapping/horse/comparing two horse autoencoders from different angles - keras.ipynb
    \centering
    \begin{tabular}{cc}
         \multicolumn{2}{c}{\includegraphics[width=.5\textwidth]{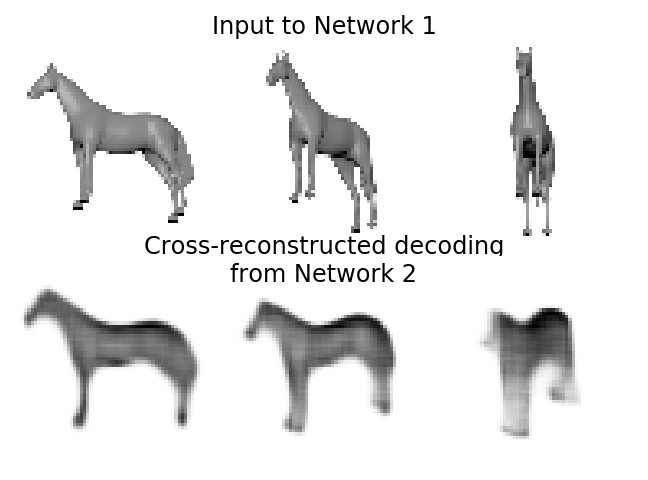}}\\
         \multicolumn{2}{c}{(a)}\\
         \includegraphics[height=.18\textheight]{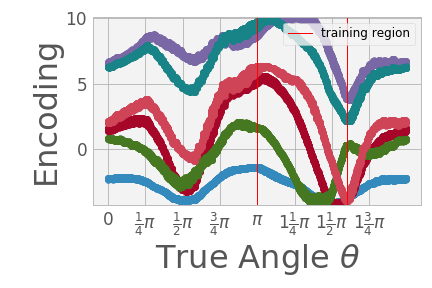}
         &
         \textbf{\includegraphics[height=.18\textheight]{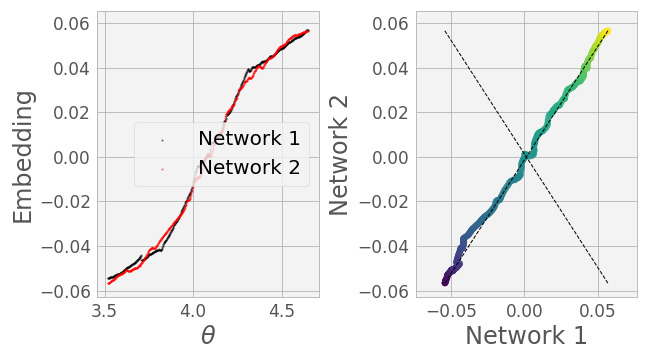}}
         \\
         (b) & (c)
    \end{tabular}
    \caption{
        \textit{Reconstructing one camera view from another.}
        We create a dataset of $3/5$ of a complete rotation of a horse figure,
        viewed from above and below, and train two convolutional autoencoders on these data.
        We then construct two Mahalanobis diffusion maps of these encodings, 
        find the smallest orthonormal transformation between these, 
        and then use this chain of transformations to construct the output of one autoencoder from the input of the other.
% \todo{we do not need the DMAP to DMAP picture anymore, only MDMAP to MDMAP. We already explain that DMAP2DMAP is bad in sec \ref{sec:differentInputs}}
        % Is the horse's head or tail closer to the camera?
        % %
        % We create a dataset of one full rotation of this figure,  
        % but then, by construction, history information form the rotation process is required to arrange the images in a circle, rather than just an arc.
        % %
        % Note that this is an additional ambiguity on top of the indeterminate direction of rotation, as seen in the spinning dancer illusion.
        % %
        % This additional ambiguity arises because the horse model, like a Necker cube but unlike the spinning dancer, is axially symmetric.
%% \todo{Regnerate all figures to be more pretty, and have better axis labels and no titles.}
    }
    \label{fig:horses}
\end{figure}

To complete the example, we compute ``cross-decodings'' by
(1) evaluating an image $x(\theta)$ from the first dataset on the first encoder, 
(2) interpolating this to its corresponding Mahalanobis diffusion map embedding,
(3) finding the nearest embedding in the other network's diffusion map,
(4) interpolating this to the second network's encoding, and finally 
(5) evaluating the second network's decoder with this encoding.
The result is shown in Fig.~\ref{fig:horses}.
When the first image is rotated by sweeping $\theta$ through the training range
(not shown here),
the reconstructed second view rotates correspondingly.
Note that the map from one embedding to the other
is an orthogonal transformation (here, simply multiplication by $+{1}$ or ${-1}$),
and can, in general, be computed as an unscaled SVD $U \cdot V^T$,
using only a small set of correspondences (here, just two pairs of corresponding points are enough to determine the sign).
This is what allowed us to perform Step 3 in ``cross-decoding" above.

%\comment{Did I leave anything out here?}
%\todo{Make this less verbose. I think that a table, however, would just take up more space.}

\section{Discussion}

In this work we formulated and implemented algorithms that construct transformation functions between observations of different deep neural networks in an attempt to establish whether these networks embodied realizations of the same phenomenon or model.
We considered observations of the activations of both output neurons and internal neurons of the networks; we assumed these are smooth (nonconstant!) functions over an input manifold. This allowed us to employ Whitney's theorem as an upper bound on how many of these functions are needed to preserve the input manifold topology. With the transformation, we could map from the input of one network, through its activations and the transformation, to the state (activations) and output of the other network. We also explored the generalization properties of the networks and how our transformation fails---as we explore input space---when the observations cannot be mapped to each other any more. 
Our transformations go beyond the probabilistic correspondences typically imposed by VAEs, towards point-wise correspondences between manifolds.

Open challenges: {\bf (1)}  Our approach hinges on being able to observe not just a single input processed by the network, but also how a small neighborhood of this input is processed by the network. Obtaining such neighborhoods on the input manifold in a way consistent across the two networks is a nontrivial component of the work---it constitutes part of our {\em observation process}, of the way data used for the training of the two networks have been collected. Our work suggests that protocols for data collection that provide such additional information can be truly beneficial in model building~\citep{moosmueller-2020,dietrich-2020}; 
% This may be a challenge, albeit more simple than in experiments, because we can use automatic differentiation of the neural network. This is not possible if the input manifold is not a simple linear space, because the covariances then would not be with respect to the manifold geometry but its ambient space.
{\bf (2)} Which internal neurons to pick for the input space embedding? In principle, any $2d+1$ neurons would be enough, but in practice, different choices vary widely in the curvature they induce on the embedding, leading to different numerical conditioning of the problem;
{\bf (3)} How to deal with intrinsically high-dimensional input manifolds, or widely varying sampling densities? These would necessitate large training data sets. Recent work on Local Conformal Autoencoders~\citep{peterfreund-2020} and their empirically observed good extrapolation performance may prove helpful.

\section{Broader Impact}

%\todo{Authors are required to include a statement of the broader impact of their work, including its ethical aspects and future societal consequences. }

The construction of transformation functions between different neural networks (and different data-driven, possibly even physically informed, models in general) has wide-ranging implications, since it enables us to calibrate different data-driven models to each other. 
It also holds the promise of allowing us to improve qualitatively correct (but quantitatively inaccurate) models by calibrating them to experimental data.
Starting with one model and calibrating it to another is, in effect, a form of transfer learning and domain adaptation.

It may be possible to calibrate different models to each other over a large portion of input space, yet the calibration may fail for inputs far away from the training set. Our procedure allows us to explore the way this calibration (the ability to qualitatively generalize) fails, {\em by locating singularities in the transformation} as we move away from the training set, exploring input space. Systematically exploring the nature and onset of these singularities, and what they imply about the nature of the underlying physics is, we believe,
an important frontier for data-driven modeling research.
%
% In order to facilitate reuse of these methods, 
% the data and Python 3 code have been made available on a public code-hosting respository
% at \texttt{ADDRESS ADDED AFTER REVIEW}.

% Authors should discuss both positive and negative outcomes, if any. For instance, authors should discuss a)  who may benefit from this research, b) who may be put at disadvantage from this research, c) what are the consequences of failure of the system, and d) whether the task/method leverages biases in the data. If authors believe this is not applicable to them, authors can simply state this.

% Use unnumbered first level headings for this section, which should go at the end of the paper. {\bf Note that this section does not count towards the eight pages of content that are allowed.}

\begin{ack}

This work was partially supported by the DARPA PAI program,
as well as
the U.S. Army Research Laboratory,
and the U.S. Army Research Office under contract/grant number W911NF1710306.
F.D. would also like to thank the Department of Informatics at the Technical University of Munich for their support. It is a pleasure to acknowledge discussions with Drs. J. Bello-Rivas and J. Gimlett.

\end{ack}

\medskip

\small

\bibliographystyle{plainnat}
\bibliography{literature_koopman,generativeModeling,neuralnetworks}

\end{document}